\documentclass{article}
\pdfoutput=1
\usepackage[latin1]{inputenc}
\usepackage[T1]{fontenc}
\usepackage{authblk}
\usepackage{pstricks}
\usepackage{pst-node}
\usepackage{pst-plot}
\usepackage{graphicx}

\title{Automated Control and Simulation of Dynamic Robot Teams in the Domain of CFK Production}
\author[1]{Marian K\"orber}
\author[2]{Roland Gl\"uck}
\affil[1,2]{Deutsches Zentrum f\"ur Luft- und Raumfahrt}
\affil[1]{marian.koerber@dlr.de}
\affil[2]{roland.glueck@dlr.de}
\date{}

\begin{document}
\maketitle

\section{Introduction}

Pick and place processes in aerospace industry have become an area of research interest during the last decade, see e.g.~\cite{BjoernssonReview,AngererAutomationSystem,EckardtDryFiber}. Also, generation of programs for robots and controlling robots automatically from data of the robot cell and CAD data of the tooling has already a long history.
In the domain of CFK production for aircraft parts previous work includes~\cite{SchusterKOKO,SmartManufactoring2018,TailorMadeThermoplastics2017,AutomatedProductionCAMX2017,FlexibleThermoplasticsCAMX2017}.
The main idea is always to deduce grip and drop positions of grippers in a pick and place process of CFK plies from a CAD file (enhanced with additional informations about the transformation between plies in 2D at the gripping position and in 3D at the dropping position) and to us these positions to generate trajectories for the production process.
In constrast to this work, we provide new ideas and techniques dealing with the following issues:
\begin{itemize}
 \item order of the laying steps: in all of the cited work, the order of the laying steps was not considered at all or it was derived from the linear order in the CAD file. 
 We free ourselves from this rigid specification and determine only a partial order from the geometric properties of the CAD file.
 \item parallel execution: in contrast to the cited work, we also allow parallel execution of (sub)tasks.
 \item flexible assignement: while the cited work operates with unchanging one-to-one assigments of grippers to plies we allow multiple gripper configurations for one ply.
\end{itemize}
All these extensions allow a more flexible schedule and open the door for optimized process executions.

The approach we follow is closely related to the one described in~\cite{ICINCOTeamBots} where also the use case was sketched. It consists of the four phases analysis, assignment, scheduling and planning, and simulation or execution. The analysis and assignment phases correspond to Section~\ref{sec:modellingAndPrecomputation}, the scheduling and planning phase to Section~\ref{sec:approaches}.

\section{Set Up}\label{sec:setUp}
\subsection{Plybook}
\begin{figure}[h]
\begin{center}
\includegraphics[width=0.8\textwidth]{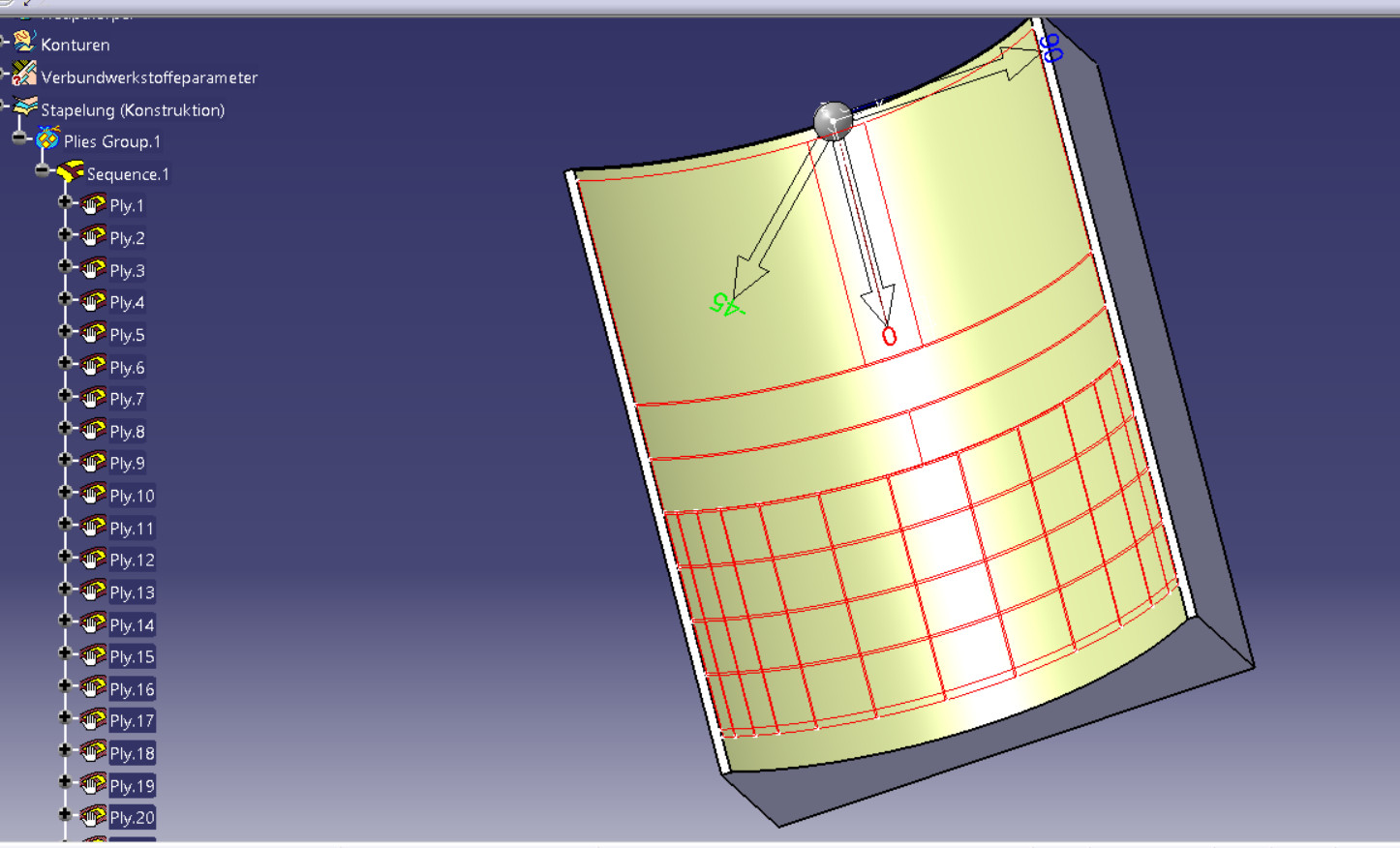}
\end{center}
\caption{The Example Plybook}
\label{fig:CATIA}
\end{figure}

The plybook was created in CATIA, a standard tool for construction in the area of CFK.
For our demonstrating example we used a semicylindric tooling as mock-up emulating a aircraft fuselage as used also in real production.
The lay-up consists of alltogether 23 carbon plies of different size und form which are partially overlapping.
A screenshot from the construction is given in Figure~\ref{fig:CATIA}.

\subsection{Cell Layout}
%Beschreibung der Zelle und der Greifer ..
As production cell we created a cell similar to the facilities at the DRL in Augsburg.
It consists of two upright KUKA robots mounted on two parallel linear axes.
Both of them are equipped with two identical grippers, called snake grippers (``Schlangengreifer'') which are real existing grippers at the DLR used already in practice.
These grippers have vacuum suction devices on their bottom which can grasp carbon fiber textiles and even metal sheets.
\begin{figure}
\begin{center}
\includegraphics[width=0.8\textwidth]{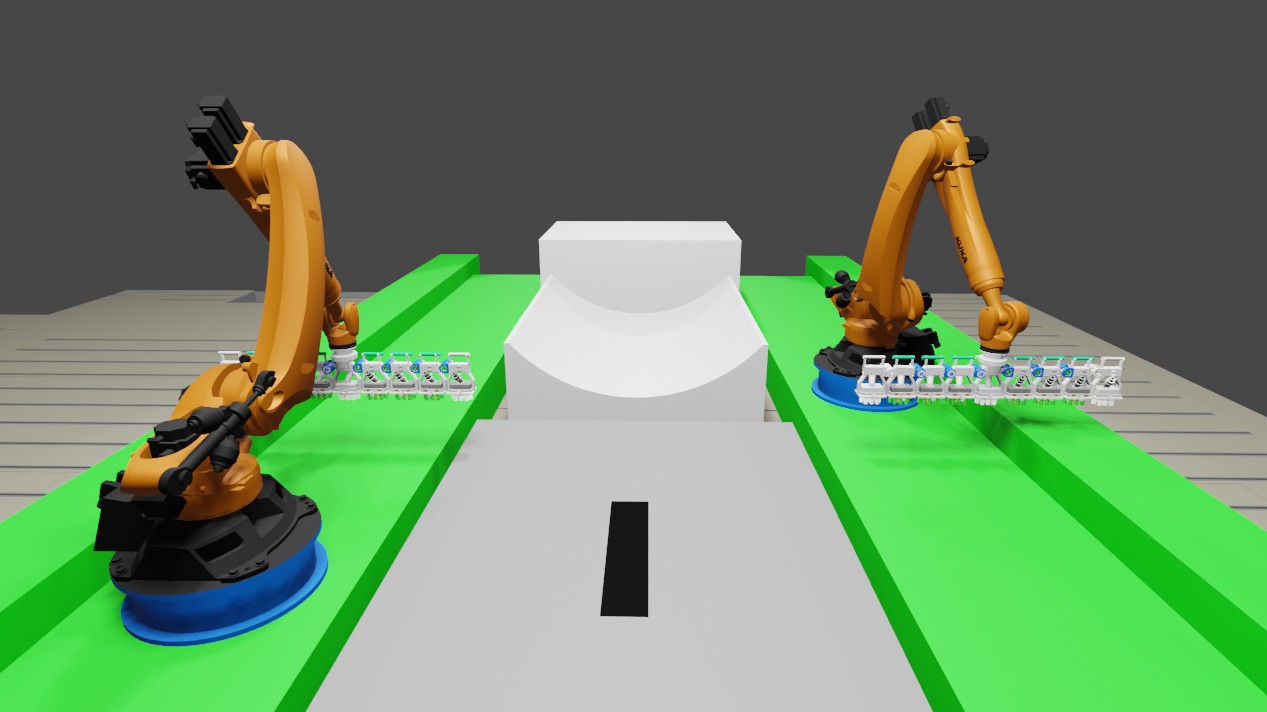}
\end{center}
\caption{The Cell Layout}
\label{fig:cell}
\end{figure}
Figure~\ref{fig:cell} shows this scenario with the robots at the right and the left with the grippers mounted on them, the desk from where the plies are picked up with an already placed ply in the front, and the tooling form in the background.

\section{Modelling and Precomputation}\label{sec:modellingAndPrecomputation}
In order to use automated planning tools as described in section~\ref{sec:approaches} we have to some modelling and precomputation. The main purpose of the present work is not offline planning of pick and place processes but rather the decomposition and subsequent scheduling of such processes. 

One basic step in our modelling is the assignement of grippers to plies, i.e., which gripper or which gripper teams can be used to pick and place a ply under consideration. 
To assign grippers or gripper teams to plies we have to determine some geometric ply characteristics like shape, curvature and size as well as the ply's material. 
Additionally, grippers are classified by geometric attributes like width, height and their deformability, indicating whether they can be used for picking and placing plies of given shape and goal curvature. 
Another attribute of a gripper is its picking method, i.e., whether it uses vacuum gripping, volume flow or needles. 
The gripping method is associated with the material; so a ply permeable to air can not be handled with a vacuum gripper. 
In our setup, the present grippers can deal with the material from the plybook; however, our approach is expected to deal also with plybooks with varying materials since the derived actions abstract from these conditions.

Rectangular plies under a given threshold size can be handled by one single gripper; in our scenario, every gripper can handle one such ply. 
For the positioning of the gripper, we compute the smallest enclosing rectangle and position the gripper parallel two this rectangle's sides. 
For greater simply curved plies we use two such grippers, positioned at the rims of the longer side of the smallest enclosing rectangle. 
If the goal position of the ply is doubly curved, we have to use other grippers which can be positioned using already existing scripts.

Another important point concerns the order in which plies have to be placed in the form. 
Traditionally, plies are placed in the form in the same top-down order in which they appear in the plybook. 
However, this order may be random and not justified by geometric or process reasons. 
For example, two symmetrical, non-overlapping plies can be placed in any of the two possible orders, leading to a permissible process. 
To model this observation we compute a so called \emph{dependency matrix}, a two-dimensional Boolean matrix where the entry at position [i][j] indicates whether the i-th ply should be placed before the j-th ply. 
The criteria are that the i-th ply has to be placed before the j-th ply were that the i-th and the j-th ply overlap and that the j-th lies over the i-th ply in the overlapping area. 
Note that this construction may overlook transitivity, i.e., it could be the case that the first ply has to be placed before the second one by the given criteria and the second one before the third one but that the first and the third ply do not overlap. 
This can be overcome by simply computing the transitive hull of the obtained Boolean matrix. 
However, this may not be necessary since the planner tools described in section~\ref{sec:approaches} will deduce transitive dependencies on their own.

From this matrix, we compute another Boolean matrix indicating which plies can be placed immediately after each other. 
Given such a pair of plies we can optimize the overall process duration if we pick the second ply (the one which has to be placed later) with one robot while placing the first one with the other robot enabling parallization of these tasks.

\section{Approaches}\label{sec:approaches}

In this section we describe and compare different approaches for the scheduling of actions in the described scenario.

\subsection{Traditional Approach}\label{ssec:tradApproach}
The most used approach in the described scenario processes the plybook ply by ply in the order in which the plies appear in the plybook. A manual assignment of grippers to plies is accompanied by automated computation of drop points and offline trajectory planning. Examples for this approach are described in~\cite{SchusterKOKO,SmartManufactoring2018,Frommel2020470} or in the context of fibre metal laminate in~\cite{VisteinGLARE}. Here, the overall process is first partitioned in a linear sequence of steps associated to each ply which in turn are broken down to the pick, the transport and the place process. Of course, the mentioned work does not aim for optimal scheduling but deals rather with offline programming and process issues in real world applications. 

\subsection{Using PDDL}\label{ssec:usingPDDL}
In the following, we describe two approaches using the Planning Domain Definition Language (PDDL, see e.g.~\cite{PlanningWikiRef}). It was introduced 1998 by McDermott for the International Planning Competition (IPC, see~\cite{IPC}) which developed into a regular event setting benchmarks in automated planning.

PDDL models planning problems in two files, one domain file and one problem file. The domain file describes the data types occurring in the setting, introduces predicates and (in advanced versions of PDDL) numerical functions and defines actions based on the involved resources and the action's preconditions and effects. In the problem file, a concrete instantiation of the domain file is given, consisting of instances of the objects, an initial state and a goal state, both in terms of the predicates defined in the domain file. Also, advanced versions of PDDL allow a metric which can be used to minimize the overall duration of the process.

There is an enormous list of planners for solving problems defined in PDDL, see e.g.~\cite{PlanningWikiPlanners} for an alphabetically ordered list. 
To find a planner suitable for our problem we had to look for a solver capable of dealing with metrics a described above. 
A solver fulfilling all these criteria is the ENHSP solver from~\cite{ENHSP} which we used for our planning.

The overall workflow for this approach is depicted on Figure~\ref{fig:worklflow}: from the CAD File we extract first the plies and their properties like shape, size and curvature. Together with the list of gripper properties we determine which gripper or which gripper teams can handle which ply. Together with the cell layout which contains information about the available gripper and gripper teams we construct a PDDL formulation of our problem which is solved subsequently. This solution is transformed back (using information from the CAD-file and the cell layout) to concrete robot motions which are fed into Blender fro simulation.

\setlength{\unitlength}{0.9cm}
\begin{figure}
\begin{picture}(10, 10)(-4,-2)
\put(0,9){\framebox(3,1){CAD-File}}
\put(1.5,9){\vector(0,-1){1}}
\put(0,7){\framebox(3,1){Ply Properties}}
\put(1.5,7){\vector(0,-1){1}}
\put(0,5){\framebox(3,1){
\begin{tabular}{c}
     Gripper  \\
     Assignement 
\end{tabular}
}}
\put(4,5){\framebox(3,1){
\begin{tabular}{c}
     Gripper \\
     Properties 
\end{tabular}
}}
\put(4,5.5){\vector(-1,0){1}}
\put(1.5,5){\vector(0,-1){1}}
\put(0,3){\framebox(3,1){
\begin{tabular}{c}
     PDDL \\
     Problem
\end{tabular}
}}
\put(4,3){\framebox(3,1){
\begin{tabular}{c}
     Cell \\
     Layout 
\end{tabular}
}}
\put(4,3.5){\vector(-1,0){1}}
\put(1.5,3){\vector(0,-1){1}}
\put(0,1){\framebox(3,1){PDDL Solution}}
\put(4,1){\framebox(3,1){
\begin{tabular}{c}
     Back-\\
     transformation 
\end{tabular}
}}
\put(4,1.5){\vector(-1,0){1}}
\put(1.5,1){\vector(0,-1){1}}
\put(0,-1){\framebox(3,1){
\begin{tabular}{c}
     Simulation \\
\end{tabular}
}}
\end{picture}
\caption{Workflow of the PDDL approach}
\label{fig:worklflow}
\end{figure}
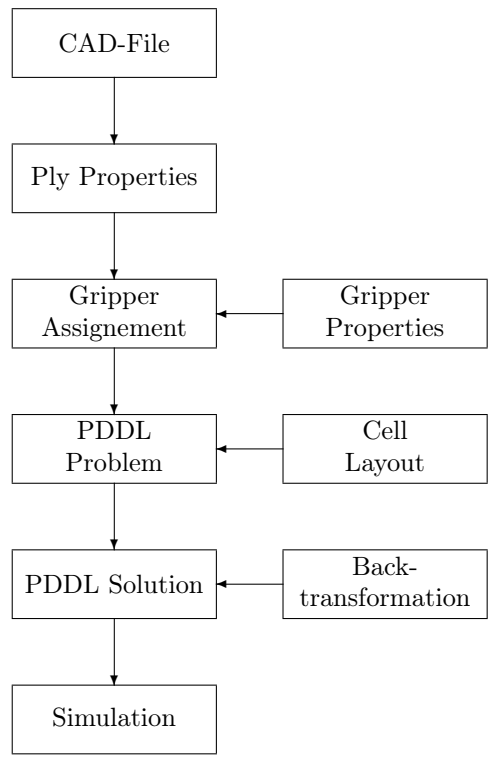

\subsubsection{Simple Approach}\label{sssec:simpleApproach} The first and rather raw approach is basically a PDDL version of the ideas described in~\ref{ssec:tradApproach}. The atomic actions are here the picking of a ply, the placing of a ply (including the transport) and the movement of a robot back to the table. In this case, we had as actions only the picking and placing of a ply and the drive of the robot at the table. Depending on the robots associated with each ply as described in section~\ref{sec:modellingAndPrecomputation} we have the action that one robot picks and drops a ply or that both robots have to fulfill these tasks. Additionally, each action was assigned a duration increasing the overall duration after its execution. We omit the condition that a ply is at the table because we focus on the pick and place process and a ply can be delivered at the table at random by a suitable conveyor system without affecting the rest of the setup.

In the problem file, we defined the initial state where both robots stand at the table, no ply is placed in the form and the overall duration equals zero. In the goal state, every ply has to be placed at its place in the form and the robots have to be back at the table. Of course, we demanded in the metric part the minimization of the overall duration. 

\subsubsection{Using Parallelization}\label{sssec:usingParallelization}
This approach takes into account the possibilities of parallel execution of tasks. The problem file is the same as described above but the domain file contains additional possible actions. In particular, we introduce the following actions modelling parallel executions:

\begin{itemize}
    \item parallel picking of one and placing of another ply
    \item parallel placing of a ply by one robot and drive at the table of the other robot
    \item parallel picking of a ply by one robot and drive at the table of the other robot
    \item parallel drive at the table of both robots
\end{itemize}

Note that there is no parallel picking or placing of two plies. Parallel picking is impossible since we stipulate that there is always at most one ply on the table. The modelling of parallel placing is not necessary because as a precondition for parallel placing we would have two robots, each of them holding already a ply. Since parallel picking was ruled out, one robot has to remain at the table during the picking process of the other robot. However, in our modelling, the robot already holding a ply could perform a placing action parallel to the picking action of the other robot.

%\subsection{Ludwigs Approach}
%Ansatz aus der Diss von Ludwig --

\section{Simulation Environment}
%Beschreibung der Simulationsumgebung --
To visualize our results evaluate them virtually we had to look for a suitable simulation environment.
We identified easy handling of robot motions, a powerful physics engine for the simulation of the textile plies and a state of the art rendering machine for the viualization as necessary conditions for this environment.
Eventualls we chose Blender (see~\cite{Blender}) which has also the advantage of being licensed in GPL.
An impression of it can be seen in the picture from Figure~\ref{fig:cell}.
The final output of the simulation is an .avi-file which shows the process where one can position also the camera as desired and has also full control over Blender's rendering machinery.

For the actual simulation the results from Section~\ref{sec:approaches} were exported as JSON containing the actions to be taken by each robot and  spatial informations about the ply and the gripping and dropping positions, resp.
A library of Python scripts was created which parsed these results and controlled the robots and generated the adhesion forces for the grippers.
The informations about the contour of ply were only used for the gripping process, the transport and the dropping was left to Blender's physics engine.
This leads to a reasonably realistic behavior of the material; see also the remarks in Section~\ref{sec:results}.

\section{Results}\label{sec:results}
%Vergleich der verschiedenen Ansätze, natürlich alles besser als der traditionelle Ansatz --
As one would expect the approach described in Part~\ref{sssec:usingParallelization} outperformed the other approaches concerning overall process time.
This has to be seen with the caveat of the immediate furnishing of plies as already mentioned in Part~\ref{sssec:simpleApproach}.

Other, also insightfull conclusions could be drawn from the simulation (and can be useful in future simulations):
\begin{itemize}
 \item Collisions between the involved robots, desks and forms can be detected and dealt with.
 \item Due to the physics engine, the behavior of the material can be predicted to a certain extent. 
 This allows to identify situations where the ply may be sag too much if the grippers are too close to each other.
 Conversly, situations where the grippers are directed too far apart lead to a breakaway of the material which also can be seen in the simulation.
\end{itemize}

\section{Outlook}
%-- Einsatz im realen Leben, AI, Quantencomputing, ... --
This work showed a step towards automation in an area where in real world production manual work is still rampant.
Also, planning and scheduling of automated tasks in this domain is still done in large parts by hand which is time expensive and may often lead to sub-optimal solutions.
We plan to evaluate the presented approach on a real world test case at the DLR facilities in Augsburg.
This may also help to find mistakes in the planning and scheduling modules.

The planning was demonstrated here for a special class of grippers and geometries.
Is is applicable to other grippers of a comparable, i.e., single curved and longish, grippers, regardless of the gripping method.
For other srippers and geometries, especially when faced with doubly curved geometries, other gripping algorithms have to be integrated.
The DLR developed already methods for pick and place processes involving such situations; the integration in this framework is possible without problems.
The greates remaining amount of work will be the modelling in Blender.

A completely different direction of future work concernes the scheduling mechanism.
PDDL solvers can not produce a provably optimal solution due to the state explosion problem but have powerful tools for generating nearly optimal solutions.
Methods taken from AI could lead in our special case to better results for this part of our workflow.

\textbf{Acknowledgements: } This work was partially funded by the DFG via the project ``TeamBots'' for what we are very grateful.

\bibliographystyle{plain}
\bibliography{main}
\end{document}